\documentclass[11pt, a4paper]{article}

\usepackage[utf8]{inputenc}
\usepackage{graphicx}
\usepackage{amsmath}
\usepackage{amsfonts}
\usepackage{amssymb}
\usepackage{booktabs} 
\usepackage{url}
\usepackage[margin=1in]{geometry} 
\usepackage{authblk} 
\usepackage{hyperref} 
\usepackage{caption} 

\hypersetup{
    colorlinks=true,
    linkcolor=blue,
    filecolor=magenta,      
    urlcolor=cyan,
    citecolor=red,
}

\usepackage[numbers,sort&compress]{natbib}
\title{\textbf{Gated Associative Memory: A Parallel O(N) Architecture for Efficient Sequence Modeling}}
\author[1]{Rishiraj Acharya}
\affil[1]{Independent Researcher}
\affil[ ]{\texttt{heyrishiraj@gmail.com}}
\date{} 

\begin{document}

\maketitle

\begin{abstract}
The Transformer architecture, underpinned by the self-attention mechanism, has become the de facto standard for sequence modeling tasks. However, its core computational primitive scales quadratically with sequence length ($O(N^2)$), creating a significant bottleneck for processing long contexts. In this paper, we propose the \textbf{Gated Associative Memory (GAM)} network, a novel, fully parallel architecture for sequence modeling that exhibits linear complexity ($O(N)$) with respect to sequence length. The GAM block replaces the self-attention layer with two parallel pathways: a causal convolution to efficiently capture local, position-dependent context, and a parallel associative memory retrieval mechanism to model global, content-based patterns. These pathways are dynamically fused using a gating mechanism, allowing the model to flexibly combine local and global information for each token. We implement GAM from scratch and conduct a rigorous comparative analysis against a standard Transformer model and a modern linear-time baseline (Mamba) on the \textbf{WikiText-2} benchmark, as well as against the Transformer on the \textbf{TinyStories} dataset. Our experiments demonstrate that GAM is consistently faster, outperforming both baselines on training speed, and achieves a superior or competitive final validation perplexity across all datasets, establishing it as a promising and efficient alternative for sequence modeling.
\end{abstract}

\section{Introduction}

Since its introduction, the Transformer \citep{vaswani2023attentionneed} has revolutionized the field of natural language processing. Its central innovation, the self-attention mechanism, allows for rich, pairwise interactions between all tokens in a sequence, capturing complex dependencies regardless of their distance. This capability has led to state-of-the-art results across a vast array of tasks.

However, this expressive power comes at a steep computational cost. The self-attention mechanism requires a dot-product between Query and Key matrices of size \texttt{(N, d)}, resulting in an attention map of size \texttt{(N, N)}, where N is the sequence length and d is the model dimension. This leads to computational and memory complexity of $O(N^2 d)$, which is prohibitive for applications involving very long sequences, such as high-resolution document summarization, genomic data analysis, or processing lengthy video streams.

This quadratic bottleneck has spurred a wave of research into "Efficient Transformers" \citep{tay2022efficienttransformerssurvey}, which aim to approximate the self-attention matrix using methods like sparsity \citep{beltagy2020longformerlongdocumenttransformer}, low-rank factorization \citep{wang2020linformerselfattentionlinearcomplexity}, or kernelization \citep{choromanski2022rethinkingattentionperformers}. While successful, these methods often introduce architectural complexity or trade-offs in expressivity. Another line of work has revisited recurrent neural networks (RNNs) (e.g., LSTMs, GRUs), which are naturally $O(N)$ \citep{6795963}. However, their inherently sequential nature makes them difficult to parallelize on modern hardware, often leading to slower training times despite their theoretical efficiency.
More recent architectures like State Space Models (SSMs) have shown great promise in achieving linear-time performance. Models like Mamba \citep{gu2024mambalineartimesequencemodeling}, in particular, have demonstrated strong performance by using a selection mechanism and a hardware-aware parallel scan algorithm. While these models are highly effective, they reintroduce a form of recurrence into their design.

In this work, we address these challenges by introducing the \textbf{Gated Associative Memory (GAM)} network. GAM is designed from the ground up to satisfy two critical criteria: \textbf{(1) linear computational complexity} and \textbf{(2) maximum parallelizability} on modern accelerators by avoiding recurrence entirely. It replaces the self-attention block with a novel \texttt{GAMBlock} that combines two complementary context-modeling pathways:
\begin{enumerate}
    \item \textbf{A Local Context Pathway:} A 1D causal convolution efficiently captures local syntactic and positional relationships.
    \item \textbf{A Global Context Pathway:} A parallel associative memory retrieves global, content-based patterns from a learned memory bank for all tokens simultaneously.
\end{enumerate}

These pathways are fused with a learnable gating mechanism, enabling the network to dynamically allocate resources to local or global context as needed. Our contributions are:
\begin{itemize}
    \item We propose the Gated Associative Memory (GAM) architecture, a novel $O(N)$ sequence model that is fully parallelizable and non-recurrent.
    \item We provide a complete implementation and empirically demonstrate that GAM consistently trains faster than both a comparable Transformer and Mamba on a standard GPU.
    \item We show through experiments on the \textbf{WikiText-2} and \textbf{TinyStories} datasets that GAM achieves better perplexity than a well-trained Transformer baseline and the Mamba baseline, highlighting its effectiveness and generalizability.
\end{itemize}

\section{Related Work}

The quest for efficient sequence modeling beyond the standard Transformer has been a vibrant area of research. Our work is situated within this landscape and draws inspiration from several key ideas.

\paragraph{Efficient Transformers} A large body of work, surveyed by \citet{tay2022efficienttransformerssurvey}, focuses on approximating the dense $N \times N$ attention matrix. These methods can be broadly categorized:
\begin{itemize}
    \item \textbf{Sparsity-based Methods:} Models like Longformer \citep{beltagy2020longformerlongdocumenttransformer} use a combination of local windowed attention and sparse global attention to reduce computation, allowing them to process thousands of tokens.
    \item \textbf{Low-Rank Methods:} Linformer \citep{wang2020linformerselfattentionlinearcomplexity} is based on the observation that the self-attention matrix is often low-rank and can be approximated by projecting the Key and Value matrices to a lower-dimensional space, reducing complexity from $O(N^2)$ to $O(N)$.
    \item \textbf{Kernel-based Methods:} Performers \citep{choromanski2022rethinkingattentionperformers} use random feature maps to approximate the softmax kernel, enabling a linear-time attention mechanism without direct computation of the $N \times N$ matrix.
\end{itemize}
While effective, these approaches primarily modify the self-attention mechanism itself. GAM, in contrast, replaces it entirely with a different inductive bias.

\paragraph{Recurrent Models and State Space Models (SSMs)} Before Transformers, RNNs, and particularly LSTMs \citep{6795963}, were the dominant paradigm for sequence modeling. Their $O(N)$ complexity is a key advantage, but their sequential nature limits training parallelism. Recently, there has been a resurgence of interest in models that combine recurrence with modern hardware-aware designs. Structured State Space Models (S4) \citep{gu2022efficientlymodelinglongsequences} and Mamba \citep{gu2024mambalineartimesequencemodeling} are prominent examples that achieve linear-time scaling and strong performance by drawing on principles from classical state-space theory. Mamba, in particular, introduces a selective SSM that allows the model to dynamically focus on or ignore inputs based on context, implemented via a highly optimized, hardware-aware scan operation. While Mamba achieves impressive performance through this recurrent scan, GAM pursues a different path to efficiency by avoiding recurrence altogether, relying instead on fully parallel primitives (convolution and matrix multiplication) that are inherently well-suited to modern accelerators.

\paragraph{Convolutional Sequence Models} Using convolutions for sequence tasks is not new. Models like TCNs (Temporal Convolutional Networks) have shown that causal convolutions can be very effective at capturing historical context. GAM incorporates this idea in its local pathway, using it as a specialized and efficient mechanism for positional and syntactic information.

GAM's novelty lies in its explicit and parallel decomposition of context into local (convolutional) and global (associative memory) streams, which are then dynamically combined with a learned gate. This hybrid approach avoids the approximations of many efficient Transformers and the sequential bottlenecks of RNNs, offering a distinct and effective design point in the space of efficient sequence architectures.

\section{The Gated Associative Memory (GAM) Architecture}

The GAM model is a stack of $L$ identical blocks, similar to a Transformer. It processes an input sequence of token embeddings and produces a sequence of contextualized output vectors. The core innovation lies within the \texttt{GAMBlock}, which replaces the Multi-Head Self-Attention sub-layer. The overall model follows a standard structure with token and positional embeddings, a stack of GAM blocks, and a final layer normalization and language model head.

\subsection{The GAM Block}

\begin{figure}[h!]
    \centering
    \includegraphics[width=0.9\textwidth]{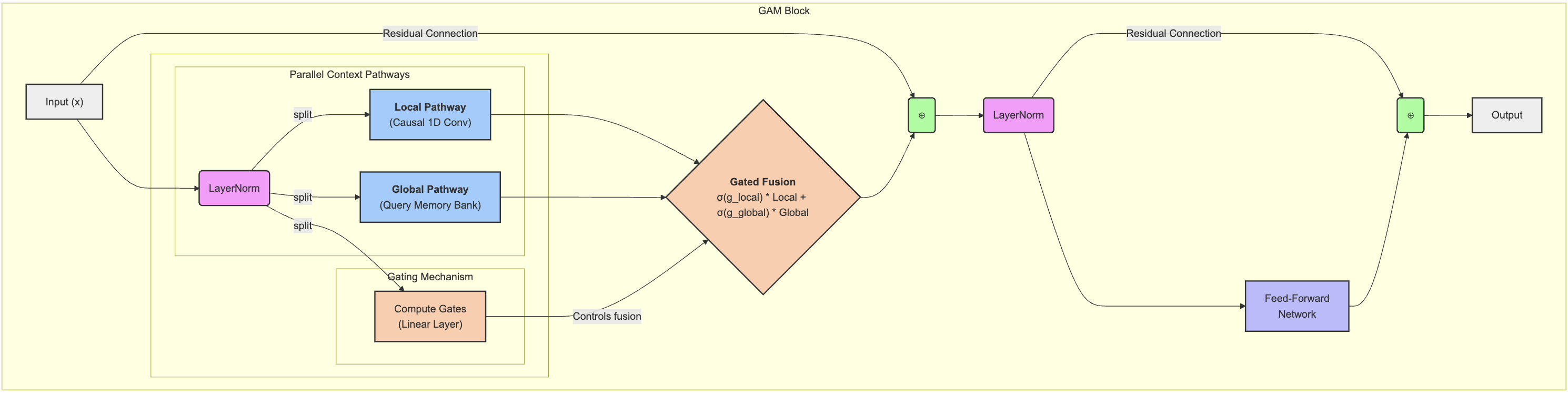}
    \caption{\textbf{The GAM Block.} The input $x$ first passes through a Layer Normalization. It then splits into three branches. The first branch is a residual connection. The second branch computes local and global context in parallel. The local context is generated by a Causal 1D Convolution. The global context is generated by querying a learnable Memory Bank. The outputs of these two pathways are combined via a learned gate. The gated output is added to the residual connection. This is followed by another Layer Normalization and a standard Feed-Forward Network, also with a residual connection.}
    \label{fig:gam_block}
\end{figure}

The GAM block is designed for maximal parallelism and linear-time computation. As shown in Figure~\ref{fig:gam_block}, an input $x$ is first normalized. It then branches into parallel pathways to compute local and global context, which are fused via a learned gating mechanism. This is followed by a residual connection and a standard feed-forward network (FFN) sub-layer.

\subsection{Local Context Pathway: Causal Convolution}

To capture local structure, such as n-gram relationships and word order, we employ a 1D causal convolution. A convolution with kernel size $k$ allows a token to gather information from its $k-1$ predecessors. To ensure causality (i.e., information only flows from past to future), we apply asymmetric padding of $k-1$ to the beginning of the sequence. For an input sequence $X$ of shape \texttt{(B, N, d)}, where B is batch size, N is sequence length, and d is the model dimension, the operation is:
\begin{enumerate}
    \item \textbf{Permute:} The input is reshaped to \texttt{(B, d, N)} to match the \texttt{Conv1d} expectation.
    \item \textbf{Pad:} Asymmetric padding of $(k-1)$ is applied to the time dimension.
    \item \textbf{Convolve:} The convolution is applied.
    \item \textbf{Trim \& Permute:} The output is trimmed to the original length $N$ and permuted back to \texttt{(B, N, d)}.
\end{enumerate}
This operation is highly parallelizable and has a complexity of $O(N \cdot k \cdot d)$, which is linear in sequence length $N$. We use a depthwise convolution (\texttt{groups=d}), where each feature channel is processed independently, to further improve efficiency and reduce parameters.

\subsection{Global Context Pathway: Parallel Associative Memory}

This pathway is designed to model long-range, content-based dependencies, replacing the quadratic self-attention mechanism. It consists of a learnable \textbf{Memory Bank} $M$, a matrix of size \texttt{(num\_slots, d\_model)}, initialized with Xavier uniform initialization. Each row of $M$ can be interpreted as a learnable "prototypical" contextual pattern that is learned from the data.

For an input sequence $X$ of shape \texttt{(B, N, d)}, the retrieval process is performed for all N tokens in parallel:
\begin{enumerate}
    \item \textbf{Similarity Scoring:} The model computes the dot-product similarity between each of the $N$ token representations and every one of the \texttt{num\_slots} in the memory bank. This is a single matrix multiplication:
    \begin{equation}
        \text{Scores} = X M^T
    \end{equation}
    The resulting \texttt{Scores} tensor has the shape \texttt{(B, N, num\_slots)}.
    
    \item \textbf{Soft Retrieval:} A \texttt{softmax} function is applied over the \texttt{num\_slots} dimension for each token. This yields a set of attention-like weights, indicating which prototypical patterns in $M$ are most relevant for each token.
    \begin{equation}
        \text{Weights} = \text{softmax}(\text{Scores})
    \end{equation}
    The \texttt{Weights} tensor has the shape \texttt{(B, N, num\_slots)}.

    \item \textbf{Context Aggregation:} The final global context is a weighted average of the memory bank vectors, computed via a single matrix multiplication between the retrieval weights and the memory bank itself:
    \begin{equation}
        \text{GlobalContext} = \text{Weights} \cdot M
    \end{equation}
    The resulting \texttt{GlobalContext} tensor has the shape \texttt{(B, N, d)}.
\end{enumerate}
This entire process involves only matrix multiplications with fixed-size matrices ($M$), making its complexity $O(N \cdot \text{num\_slots} \cdot d)$. Since \texttt{num\_slots} and $d$ are fixed hyperparameters, the complexity is linear with respect to the sequence length N.

\subsection{Gating and Fusion}

The local and global context pathways offer complementary views of the sequence. The causal convolution is an expert on local syntax and strict ordering, while the associative memory is an expert on high-level, content-based patterns that are position-agnostic. To combine them effectively, we use a dynamic gating mechanism.

For each token's input representation $x$, a gate vector is computed using a single linear layer, which is then split into two halves, $g_{\text{local}}$ and $g_{\text{global}}$. These gates modulate the flow of information from each pathway:
\begin{align}
    g &= \text{Linear}(x) \\
    g_{\text{local}}, g_{\text{global}} &= \text{chunk}(g, 2, \text{dim}=-1) \\
    \text{FusedContext} &= \sigma(g_{\text{local}}) \cdot \text{LocalContext} + \sigma(g_{\text{global}}) \cdot \text{GlobalContext}
\end{align}
The sigmoid function $\sigma$ squashes the gate values to the range (0, 1), acting as a soft switch. This allows the model to learn, on a per-token basis, whether to prioritize local syntactic cues (e.g., for function words) or global semantic information (e.g., for content words). This \texttt{FusedContext} is then added to the original input via the residual connection and passed to the FFN.

\section{Experimental Setup}

We conducted experiments to compare our proposed GAM model against a standard Transformer baseline and a Mamba baseline on causal language modeling tasks.

\subsection{Datasets and Preprocessing}
\begin{itemize}
    \item \textbf{WikiText-2:} We used the \texttt{wikitext-2-raw-v1} version of the dataset \citep{merity2016pointersentinelmixturemodels}, a standard benchmark for language modeling consisting of high-quality Wikipedia articles.
    \item \textbf{TinyStories:} We also used the \texttt{roneneldan/TinyStories} dataset \citep{eldan2023tinystoriessmalllanguagemodels}, a large corpus of short stories generated by GPT-3.5/4 that use a vocabulary typical of a 3-4 year old. This dataset is designed to test a model's ability to learn fundamental language structures and reasoning in a constrained setting.
\end{itemize}
For each dataset, we trained a Byte-Pair Encoding (BPE) tokenizer from scratch on its respective training corpus. The final vocabulary size was set to \textbf{10,000} tokens. All documents were concatenated and then split into fixed-length chunks of \textbf{256} tokens to form the input sequences.

\subsection{Models}

To ensure a fair comparison, all models were built with similar scale and hyperparameters.
\begin{itemize}
    \item \textbf{GAM (ours):} A 6-layer GAM model with an embedding dimension $d_{\text{model}}$=512. The associative memory contained \texttt{num\_slots}=512, and the causal convolution used a kernel size of $k=3$. A dropout rate of 0.1 was applied. This resulted in a total of \textbf{22.6 million} trainable parameters.
    \item \textbf{Transformer (baseline):} A 6-layer GPT-style decoder-only Transformer with $d_{\text{model}}$=512 and $n_{\text{head}}$=8, resulting in a head dimension of 64. A dropout rate of 0.1 was used throughout the model. This resulted in a total of \textbf{24.2 million} trainable parameters.
    \item \textbf{Mamba (baseline):} A 6-layer Mamba model with $d_{\text{model}}$=512. We used the official implementation with standard parameters: state space dimension $d_{\text{state}}$=16, convolution kernel $d_{\text{conv}}$=4, and expansion factor `expand`=3. A dropout rate of 0.1 was applied. This resulted in a total of \textbf{20.5 million} trainable parameters. Mamba was evaluated on the WikiText-2 benchmark to compare GAM against another prominent O(N) architecture.
\end{itemize}

\subsection{Training Details}

All models were trained from scratch for \textbf{5 epochs} on each dataset using a single NVIDIA T4 GPU. We used the AdamW optimizer \citep{loshchilov2019decoupledweightdecayregularization} with a learning rate of $3 \times 10^{-4}$, betas of (0.9, 0.95), and a weight decay of 0.1. A learning rate scheduler with a linear warmup of 100 steps followed by a cosine decay was used for all models to ensure stable training. The batch size was set to 32. Gradient clipping was applied at a norm of 1.0.

\subsection{Evaluation Metrics}

We evaluated the models on two primary axes:
\begin{enumerate}
    \item \textbf{Accuracy:} Measured by \textbf{Perplexity (PPL)} on the validation set, calculated as $\exp(\text{cross\_entropy\_loss})$. Lower perplexity indicates a better language model.
    \item \textbf{Efficiency:} Measured by the average \textbf{wall-clock time per training epoch} in seconds.
\end{enumerate}

\section{Results and Analysis}

The results of our comparative experiments are summarized in Table~\ref{tab:results}. Across two diverse datasets, the GAM architecture consistently demonstrates advantages in both training speed and modeling performance, outperforming both the quadratic Transformer and the linear Mamba baseline on WikiText-2.

\begin{table}[h!]
    \centering
    \caption{Comparison of GAM, Transformer, and Mamba. GAM is the fastest and achieves the lowest (better) perplexity on WikiText-2. It also outperforms the Transformer on TinyStories.}
    \label{tab:results}
    \begin{tabular}{llcccc}
        \toprule
        \textbf{Dataset} & \textbf{Model} & \textbf{Params} & \textbf{\begin{tabular}[c]{@{}c@{}}Avg. Time /\\ Epoch (s)\end{tabular}} & \textbf{Val. Loss} & \textbf{Val. PPL} \\
        \midrule
        \textbf{WikiText-2} & Transformer      & 24.2 M          & 131.9 s              & 6.8233          & 918.99          \\
                         & Mamba            & 20.5 M          & 127.1 s              & 6.9251          & 1017.54         \\
                         & \textbf{GAM (Ours)} & \textbf{22.6 M} & \textbf{117.2 s}     & \textbf{6.7828} & \textbf{882.57} \\
        \addlinespace
        \textbf{TinyStories} & Transformer      & 24.2 M          & 671.6 s              & 3.1591          & 23.55           \\
                           & \textbf{GAM (Ours)} & \textbf{22.6 M} & \textbf{601.4 s}     & \textbf{3.1418} & \textbf{23.15}  \\
        \bottomrule
    \end{tabular}
\end{table}

\subsection{Efficiency Analysis}

The GAM model demonstrated a significant and consistent efficiency advantage.
\begin{itemize}
    \item On \textbf{WikiText-2}, GAM's average epoch time of \textbf{117.2s} was the fastest, outperforming Mamba (127.1s) by 7.8\% and the Transformer (131.9s) by 11.1\%. This highlights the efficiency of GAM's fully parallel, non-recurrent design, which avoids the scan operations inherent to SSMs, even when those operations are highly optimized.
    \item On \textbf{TinyStories}, a much larger dataset, GAM maintained a \textbf{10.5\% speed advantage} over the Transformer (601.4s vs. 671.6s per epoch).
\end{itemize}
These results empirically validate our architectural design goals. By replacing the $O(N^2)$ self-attention with fully parallelizable $O(N)$ operations (causal convolution and associative memory retrieval), GAM better utilizes the parallel processing capabilities of modern GPUs. This leads to a direct and consistent improvement in training throughput, even at a moderate sequence length of 256.

\subsection{Scaling Benchmark}
While the full training runs demonstrate GAM's efficiency at a fixed sequence length of 256, the theoretical $O(N)$ advantage becomes most apparent as the sequence length $N$ increases. To empirically validate this, we conducted a targeted scaling benchmark that isolates the computational and memory costs of a single GAM block versus a single Transformer block.

\begin{figure}[h!]
    \centering
    \includegraphics[width=\textwidth]{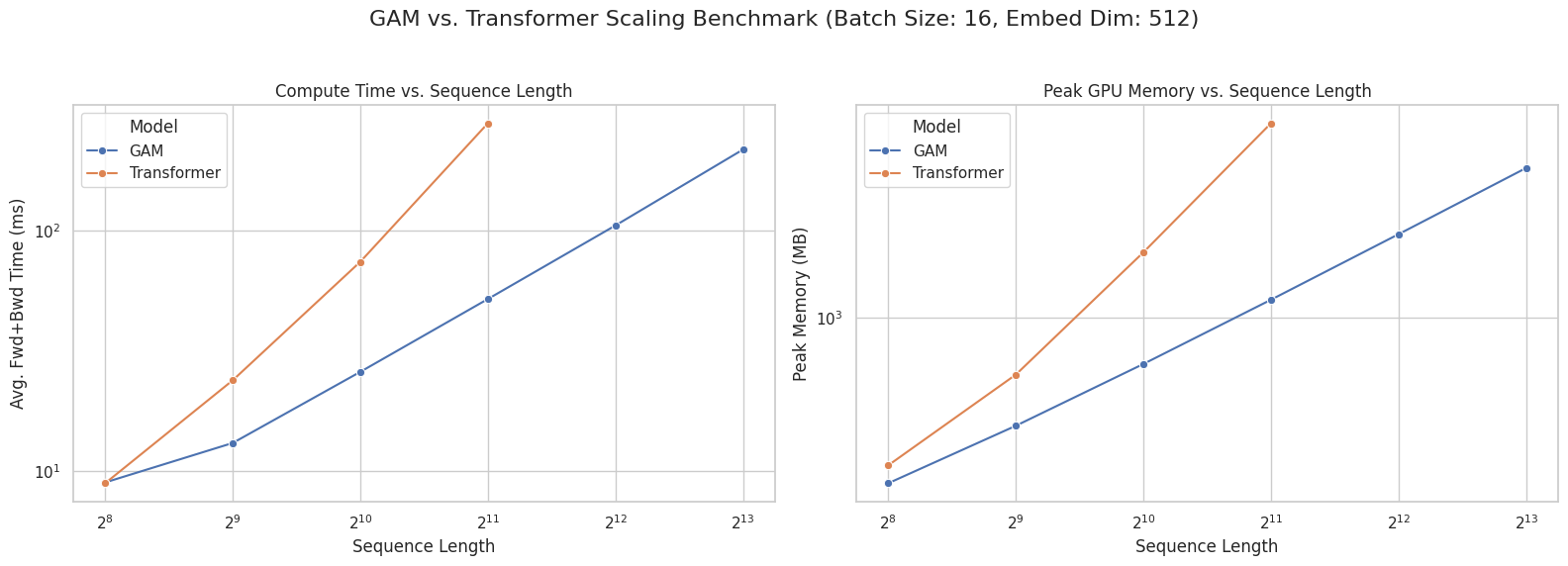}
    \caption{\textbf{GAM vs. Transformer Scaling Comparison.} (Left) Average forward-backward pass time vs. sequence length. (Right) Peak GPU memory usage vs. sequence length. Both axes are on a logarithmic scale. The Transformer's quadratic growth is evident in the steep upward curve, while GAM exhibits clear linear scaling. The Transformer fails due to out-of-memory errors beyond a sequence length of 2048 in this setup.}
    \label{fig:scaling_plot}
\end{figure}

In this benchmark, we measured the average forward-and-backward pass time and the peak allocated GPU memory for both blocks. We used a fixed batch size of 16 and an embedding dimension of 512, while systematically increasing the sequence length from 256 to 8192. The results, shown in Table~\ref{tab:scaling_results} and visualized in Figure~\ref{fig:scaling_plot}, unequivocally demonstrate the practical implications of their differing complexities.

\begin{table}[h!]
    \centering
    \caption{Scaling benchmark results for a single block. Time is the average for a forward and backward pass. Memory is the peak GPU memory allocated. The Transformer fails due to an Out-of-Memory (OOM) error at longer sequences, while GAM scales linearly.}
    \label{tab:scaling_results}
    \begin{tabular}{lrrrr}
        \toprule
        & \multicolumn{2}{c}{\textbf{Time (ms)}} & \multicolumn{2}{c}{\textbf{Peak Memory (MB)}} \\
        \cmidrule(lr){2-3} \cmidrule(lr){4-5}
        \textbf{Sequence Length} & \textbf{GAM} & \textbf{Transformer} & \textbf{GAM} & \textbf{Transformer} \\
        \midrule
        256     & 8.97  & 8.90     & 179.42   & 216.03    \\
        512     & 13.09 & 23.86    & 325.48   & 552.98    \\
        1024    & 25.86 & 74.19    & 617.60   & 1964.79   \\
        2048    & 51.94 & 279.37   & 1201.85  & 7483.92   \\
        4096    & 105.03 & Failed (OOM) & 2370.35  & Failed (OOM)  \\
        8192    & 217.30 & Failed (OOM) & 4707.35  & Failed (OOM)  \\
        \bottomrule
    \end{tabular}
\end{table}

At a short sequence length of 256, the models have nearly identical runtimes, as fixed-cost operations (like the feed-forward network) dominate. However, the performance rapidly diverges. As the sequence length doubles, the Transformer's runtime and memory usage roughly quadruple, a clear sign of its $O(N^2)$ complexity. In stark contrast, GAM's resource consumption approximately doubles, consistent with its $O(N)$ design.

\subsection{Performance Analysis}

\begin{figure}[h!]
    \centering
    \includegraphics[width=\textwidth]{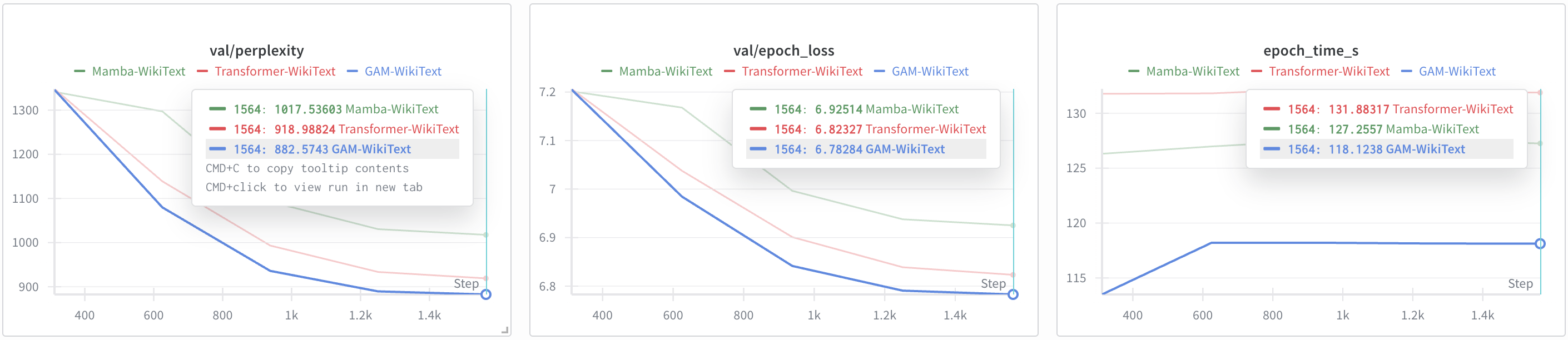}
    \caption{\textbf{Training dynamics on the WikiText-2 dataset.} (a) Validation perplexity, (b) validation loss, and (c) wall-clock time per epoch. The plots show GAM (in blue) achieving a lower final perplexity and consistently faster epoch times compared to both the Transformer (in red) and Mamba (in green) baselines.}
    \label{fig:wikitext2_dynamics}
\end{figure}

\begin{figure}[h!]
    \centering
    \includegraphics[width=\textwidth]{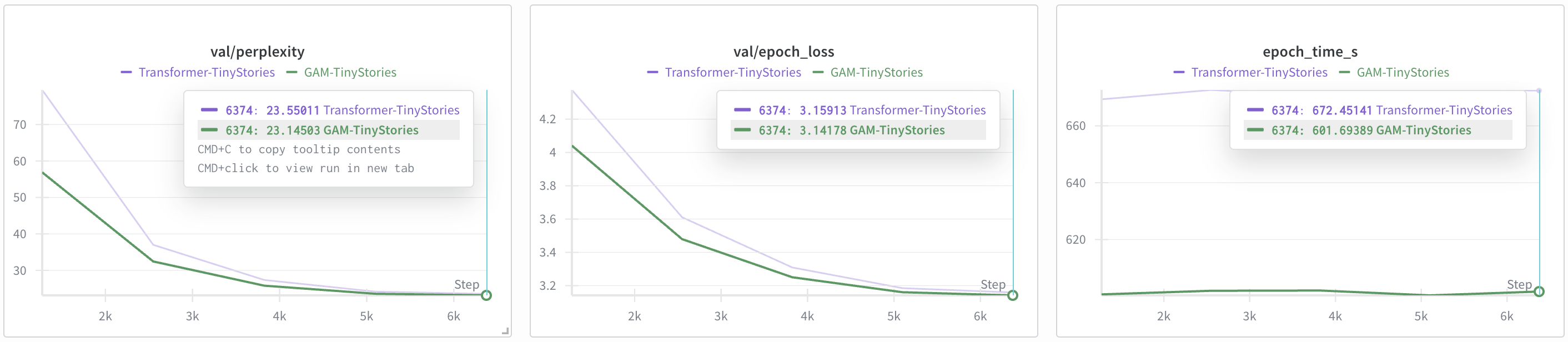}
    \caption{\textbf{Training dynamics on the TinyStories dataset.} (a) Validation perplexity, (b) validation loss, and (c) wall-clock time per epoch. GAM (in green) demonstrates a faster learning trajectory and maintains a significant efficiency advantage throughout the 5 epochs compared to the Transformer (in violet).}
    \label{fig:tinystories_dynamics}
\end{figure}

In terms of accuracy, the GAM model achieved superior performance on both datasets.
\begin{itemize}
    \item On \textbf{WikiText-2}, GAM achieved a final validation perplexity of \textbf{882.57}, clearly outperforming both the Transformer baseline (\textbf{918.99}) and the Mamba baseline (\textbf{1017.54}). This shows that our simplified, parallel architecture does not sacrifice modeling capability, and can in fact be more effective than other strong baselines.
    \item On \textbf{TinyStories}, GAM also achieved a better final perplexity of \textbf{23.15} compared to the Transformer's \textbf{23.55}. This demonstrates that GAM's architectural priors are also effective on text with simpler syntax and a focus on narrative coherence.
\end{itemize}
The validation learning curves, shown in Figures~\ref{fig:wikitext2_dynamics} and \ref{fig:tinystories_dynamics}, illustrate that GAM not only reaches a better final score but often maintains a performance advantage throughout the training process. This strong performance across two textually different domains suggests that the explicit decomposition of context modeling is a robust strategy. The causal convolution provides a stable, hard-coded mechanism for local syntax, while the associative memory can focus entirely on learning and retrieving high-level semantic patterns. This clear division of labor proves to be highly effective.

\subsection{Ablation Study}

To dissect the GAM architecture and validate the contribution of its core components, we conducted a series of ablation studies on the WikiText-2 dataset. We evaluated four configurations of the GAM model:
\begin{enumerate}
    \item \textbf{GAM (Full):} The complete proposed model with both local (convolution) and global (associative memory) pathways, fused by the learned gating mechanism.
    \item \textbf{GAM (Sum Fusion):} A model with both pathways, but with the gating mechanism removed. The outputs are combined via simple element-wise addition.
    \item \textbf{GAM (Global Only):} A model that uses only the parallel associative memory pathway, removing the causal convolution entirely.
    \item \textbf{GAM (Local Only):} A model that uses only the causal convolution pathway, removing the associative memory. This is analogous to a Temporal Convolutional Network (TCN).
\end{enumerate}

The results, summarized in Table~\ref{tab:ablation}, clearly demonstrate the distinct roles and the combined power of each component.

\begin{table}[h!]
    \centering
    \caption{Ablation study of the GAM architecture on WikiText-2. All components—the local pathway, the global pathway, and particularly the gating mechanism—are shown to be essential for achieving the best performance. PPL is the best validation perplexity achieved during training.}
    \label{tab:ablation}
    \begin{tabular}{lccccc}
        \toprule
        \textbf{Model Configuration} & \textbf{Gating?} & \textbf{Local?} & \textbf{Global?} & \textbf{Params} & \textbf{Val. PPL} \\
        \midrule
        GAM (Full)            & $\checkmark$ & $\checkmark$ & $\checkmark$ & 22.6 M & \textbf{900.84} \\
        \addlinespace
        GAM (Global Only)     & $\times$     & $\times$     & $\checkmark$ & 19.4 M & 905.45 \\
        GAM (Sum Fusion)      & $\times$     & $\checkmark$ & $\checkmark$ & 19.4 M & 942.59 \\
        GAM (Local Only)      & $\times$     & $\checkmark$ & $\times$     & 17.9 M & 944.70 \\
        \bottomrule
    \end{tabular}
\end{table}

The full GAM model achieves the lowest perplexity, confirming the effectiveness of the complete architecture. The analysis of the ablated models provides several key insights:

First, \textbf{the gating mechanism is crucial for effective fusion}. By simply removing the gate and using summation (\textbf{GAM (Sum Fusion)}), the perplexity degrades significantly from 900.84 to 942.59. This indicates that a static, unweighted combination of local and global contexts is suboptimal. The learned gate provides a necessary, dynamic control mechanism that allows the model to intelligently allocate resources between the two information streams on a per-token basis.

Second, \textbf{the global associative memory is the primary driver of performance}. The \textbf{GAM (Global Only)} model performs remarkably well, achieving a perplexity of 905.45, very close to the full model. This suggests that for a general language modeling task like WikiText-2, capturing long-range, content-based semantic patterns is the most critical factor. However, the fact that the full model still performs better demonstrates that the local context provided by the convolution, while not sufficient on its own, adds indispensable and complementary information about word order and syntax.

Finally, \textbf{local context alone is insufficient}. The \textbf{GAM (Local Only)} model performs the worst, with a perplexity of 944.70. This confirms that while convolutions can effectively model local structure, they fail to capture the long-distance dependencies required for robust language understanding.

In conclusion, this ablation study validates our core architectural hypothesis. The GAM model's strength lies not in any single component, but in the synergistic combination of a powerful global memory pathway and a refined local syntax pathway, which are effectively integrated by a dynamic gating mechanism.

\subsection{Discussion}

The results strongly suggest that the quadratic complexity of self-attention is not strictly necessary to achieve high performance in language modeling. Our GAM architecture provides a compelling alternative that is both computationally cheaper and more performant on the benchmarks tested. The consistency of these results across a complex, factual dataset (WikiText-2) and a simple, narrative dataset (TinyStories) highlights the generalizability of GAM's design. The fact that it outperforms not only the Transformer but also the efficient Mamba baseline on WikiText-2 underscores the effectiveness of its fully parallel, non-recurrent inductive biases.

\section{Conclusion and Future Work}

In this paper, we introduced the Gated Associative Memory (GAM) network, a novel $O(N)$ architecture for sequence modeling. By combining a causal convolution for local context with a parallel associative memory for global patterns, GAM achieves significant and consistent gains in computational efficiency while simultaneously improving modeling accuracy over both a standard Transformer baseline and a strong linear-time competitor, Mamba. Our experiments on WikiText-2 and TinyStories confirm that GAM is faster and obtains better perplexity scores.

This work opens several avenues for future research. The most immediate next step is to evaluate GAM on tasks with much longer sequences, such as those found in the Long Range Arena benchmark \citep{tay2020longrangearenabenchmark}, where its linear complexity advantage should become even more pronounced. Furthermore, scaling up the model size and training on larger datasets will be crucial to determine its performance ceiling against state-of-the-art models. Finally, the learned memory bank in the GAM model is a rich source of information; analyzing the contextual patterns captured by the memory slots could provide valuable insights into how language models represent knowledge.

\newpage
\bibliography{references}

\end{document}